\documentclass{article}
\usepackage{log_2022}           

\usepackage{booktabs}            % professional-quality tables
\usepackage{multirow}            % tabular cells spanning multiple rows
\usepackage{amsfonts}            % blackboard math symbols
\usepackage{graphicx}            % figures
\usepackage{duckuments}          % sample images
\usepackage{xcolor}

\usepackage[numbers,compress,sort]{natbib}
\title[Enhancing Molecular Design through Graph-based Topological Reinforcement Learning]{Enhancing Molecular Design through Graph-based Topological Reinforcement Learning}

\author[X. Zhang.]{%
Xiangyu Zhang\\
\institute{Johns Hopkins University, Baltimore}\\
\email{xzhan344@jh.edu}
}

\begin{document}

\maketitle

\begin{abstract}
% Abstracts should be a single paragraph, ideally between 4--6 sentences long.

The generation of drug-like molecules is crucial for drug design. Existing reinforcement learning (RL) methods often overlook structural information. However, feature engineering-based methods usually merely focus on binding affinity prediction without substantial molecular modification. To address this, we present Graph-based Topological Reinforcement Learning (GraphTRL), which integrates both chemical and structural data for improved molecular generation. GraphTRL leverages multiscale weighted colored graphs (MWCG) and persistent homology, combined with molecular fingerprints, as the state space for RL. Evaluations show that GraphTRL outperforms existing methods in binding affinity prediction, offering a promising approach to accelerate drug discovery.

\end{abstract}

\section{Introduction}

The design of novel molecules with specified properties is a critical yet time-consuming and resource-intensive process in drug discovery. This study proposes an innovative approach to accelerate this process by leveraging reinforcement learning (RL) in conjunction with graph features and topological information. 
Reinforcement learning, a branch of artificial intelligence, has shown promise in molecular design by making decisions based on current states to maximize rewards according to defined policies. \\

Previous applications of RL in \textit{de novo} drug design have primarily focused on generating new molecules through SMILES string manipulation \cite{staahl2019deep, olivecrona2017molecular, guimaraes2017objective, putin2018reinforced}. However, these approaches often struggled with maintaining chemical validity, a crucial factor in drug development. Zhou et al. addressed this problem by introducing MolDQN, which formulates molecular modification as a Markov decision process (MDP) \cite{zhou2019optimization}. This approach defines the MDP through a set of chemically valid actions—atom addition, bond addition, and bond removal, thereby ensuring the chemical validity of the generated molecules.While MolDQN eliminates the need for pre-training on specific datasets, it overlooks structural information, which is crucial for understanding a molecule's pharmacokinetic properties (ADME) and its interactions with biological targets. 
Feature engineering-based methods are capable of extracting structural features but are limited to merely focusing on binding affinity prediction without enabling substantial molecular modification \cite{rana2022eisa,zia2024persistent}.
To address this limitation, our study introduces a novel approach that combines Multiscale Weighted Colored Graphs (MWCG) and persistent homology embedding to capture the structural information of molecules. For each SMILES representation, we utilize the RDKit package to extract atomic coordinates and construct multiscale weighted colored graphs for each atom pair. We then employ persistent images to represent the topological information of the molecule. This structural data is combined with molecular fingerprints to create a comprehensive state representation for the reinforcement learning algorithm.
By integrating these advanced graph-theoretical and topological techniques with reinforcement learning, our method aims to generate chemically valid molecules with desired properties while considering crucial structural information. This approach has the potential to enhance the efficiency and effectiveness of the drug design process.

\section{Methods}

\subsection{Multiscale Weighted Colored Graphs}

Interaction networks of large amount of biomolecular structures data are available with the development of experimental tools \cite{xia2016review}. Graph theory, with its focus on the connectivity between vertices and edges, provides an ideal framework for representing non-covalent interactions between atoms in molecules. Our choice of Multiscale Weighted Colored Graphs (MWCG) for this study is predicated on two key advantages:
\begin{enumerate}
    \item MWCG employs color-labeled atoms to create distinct subgraphs and utilizes colored edges to represent element-specific interactions. This approach allows for a detailed capture of interactions at the atomic level.
    \item MWCG incorporates a radial basis function to scale Euclidean distances, assigning the strongest weights to edges between nearest neighbors \cite{masud2023geometric}. This weighting scheme enhances the representation of local atomic environments.
\end{enumerate}

We further augment the MWCG approach by incorporating SYBYL atom types, which enable the differentiation of elements based on their bonding environment and hybridization state \cite{ballester2014does}. This integration is particularly useful for detecting modifications resulting from valid actions in our model. For instance, a C.1 atom type would transform to C.2 upon the addition of a double bond.
In our implementation, we begin with a molecular SMILES representation and extract atom types to construct atom pairs. Subsequently, we generate multiple multiscale weighted colored subgraphs for each atom pair, with graph coloring based on SYBYL atom types. This approach enables the detection and representation of changes resulting from each valid action, effectively translating these modifications into graph features within our reinforcement learning framework. 

An undirected multiscale weighted colored graph \( G \) can be denoted as a pair \( G(\mathcal{V}, E) \), where 
\begin{equation}
    \mathcal{V}=\left\{\left(\mathbf{r}_i, \alpha_i\right) \mid \mathbf{r}_i \in \mathbb{R}^3 ; \alpha_i \in \mathcal{T} ; i=1,2, \ldots, N\right\}
\end{equation}
denotes the set of vertices of \( G \), with \( N = |\mathcal{V}| \). $\mathcal{T}$ denotes the set of all relevant atom types in a given molecular dataset. The set of edges of \( G \) is denoted by 
\begin{equation}
    E = \{e_i = (v_{i_1}, v_{i_2}) \mid 1 \leq i_1 \leq N, 1 \leq i_2 \leq N\}
\end{equation}

For biomolecules dataset, $E$ represents certain covalent or noncovalent bonds among atoms in a molecule. For a given graph $G\left(\mathcal{V}, \mathcal{E}_{k k^{\prime}}\right)$, which is used to describe molecular properties, with edge defined as follow (only consider non-covalent interactions)
\begin{equation}
    \mathcal{E}_{k k^{\prime}} = \left\{
\Phi\left(\left\|\mathbf{r}_i-\mathbf{r}_j\right\| ; \eta_{k k^{\prime}}\right) \mid \alpha_i=\mathcal{T}_k, \alpha_j=\mathcal{T}_{k^{\prime}},
\right. \\
\left.
i, j=1,2, \ldots, N ; \left\|\mathbf{r}_i-\mathbf{r}_j\right\| \leq c
\right\}
\end{equation}

where $\Phi$ is the correlation kernel that characterizes distance information through probability. $\sigma$ is the overall standard deviation between two sets. % need to specify the correct meaning 
We use this way to define the binding sets of atom type $\mathcal{T}_{k}$ and $\mathcal{T}_{k'}$. 

We use generalized exponential function as the choice of radial basis function, which is particularly effective for biomolecules suggested by \cite{opron2014fast}. The function is as follows:
\begin{equation}
    \Phi_E\left(\left\|\mathbf{r}_i-\mathbf{r}_j\right\| ; \eta_{k k^{\prime}}\right)=e^{-\left(\left\|\mathbf{r}_i-\mathbf{r}_j\right\| / \eta_{k k^{\prime}}\right)^k}, \quad \kappa>0
\end{equation}

% the meaning of kappa
Therefore, for a given molecule, the multiscale weighted colored subgraphs interaction between $k$th atom type $\mathcal{T}_{k}$ and $k'$th atom type $\mathcal{T}_{k'}$ is defined as follows:

\begin{equation}
    \mu^G\left(\eta_{k k^{\prime}}\right) = \sum_i \mu_i^G\left(\eta_{k k^{\prime}}\right) = \sum_i \sum_j \Phi\left(\left\|\mathbf{r}_i - \mathbf{r}_j\right\| ; \eta_{k k^{\prime}}\right)
\end{equation}
where $\alpha_i = \mathcal{T}_k, \quad \alpha_j = \mathcal{T}_{k^{\prime}}$

\subsection{Persistent Homology}

\begin{figure}[h]
    \centering
    \includegraphics[width=0.8\linewidth]{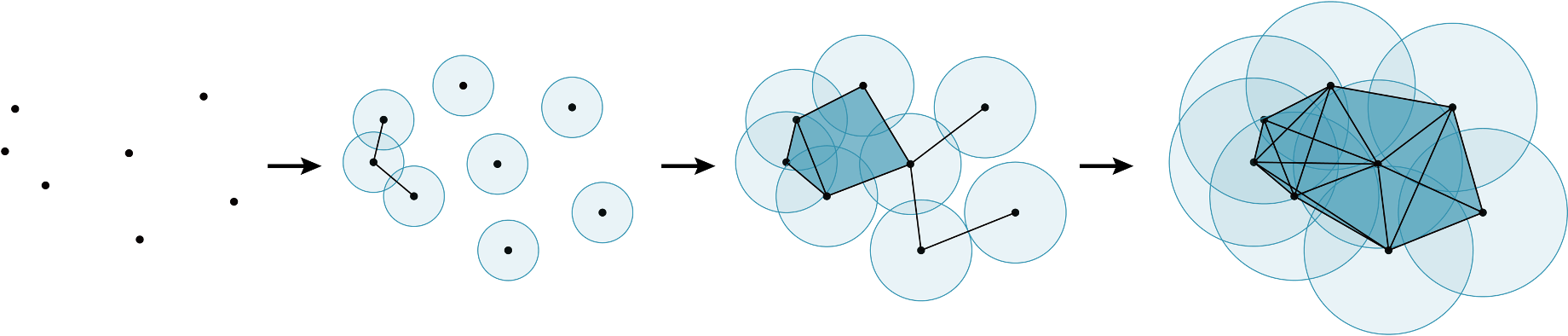}
    \caption{An Example of Filtration Process}
    \label{fig:filtration}
\end{figure}

Persistent homology is a method that utilizes the filtration process (Fig. \ref{fig:filtration}) to systematically reset the connectivity of a dataset according to a scale parameter, generating a series of topological spaces in various scales. Through filtration process, we can identify the intrinsic structure of the given molecule. 
Topological features provide several key advantages in drug design:
\begin{enumerate}
    \item They offer a robust representation of molecular shape and structure that is invariant to certain transformations, such as rotation and translation.
    \item They can capture global and local structural information simultaneously, which is essential for understanding ligand-protein interactions.
    \item Topological features can represent multi-scale structural information, allowing for the analysis of molecular properties at different levels of granularity.
\end{enumerate}

\begin{figure}[h]
    \centering
    \includegraphics[width=0.8\linewidth]{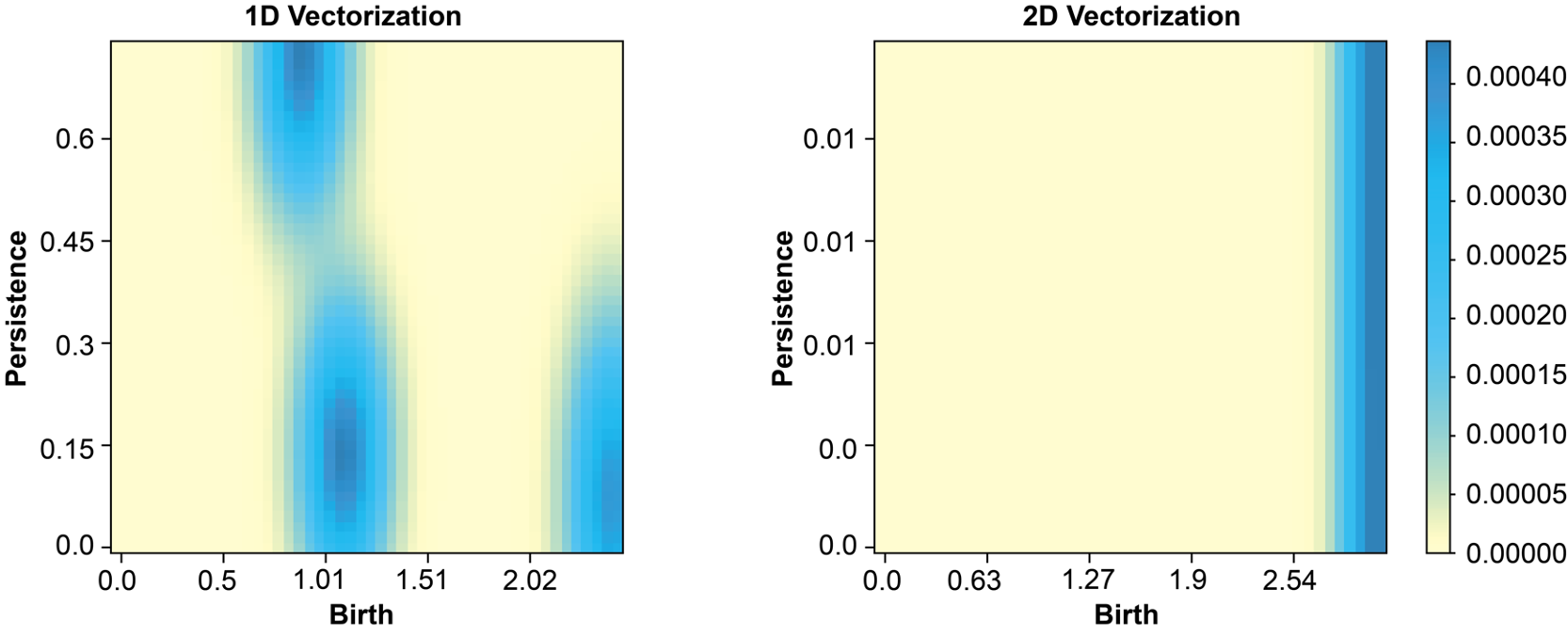}
    \caption{Example of Persistent Images}
    \label{fig: persist}
\end{figure}

To leverage these topological features within our reinforcement learning framework, we employ persistent images (Fig. \ref{fig: persist}). This approach transforms discrete topological data into continuous density distributions, making the persistent topological features more amenable to processing by reinforcement learning algorithms.

\subsection{State Space and Action Space}

\begin{figure}
    \centering
    \includegraphics[width=0.8\linewidth]{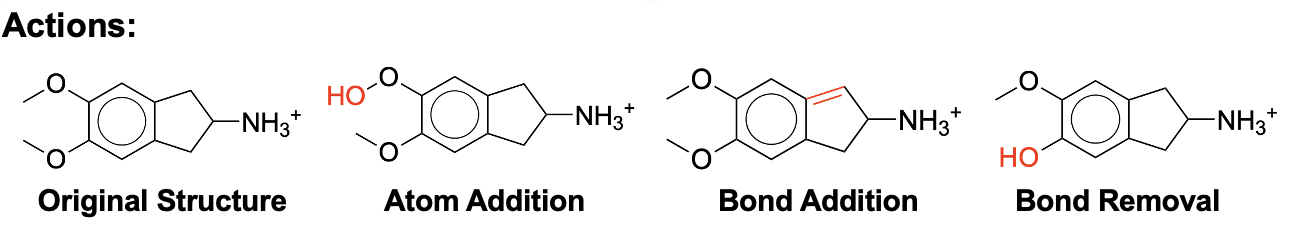}
    \caption{Visualization of Action: performing sequential actions including atom addition, bond addition, and bond removal on the structure COc1cc2c(c1OC)CC([NH3+])C2 to explore its optimization potential}
    \label{action}
\end{figure}

\begin{figure}[h]
    \centering
    \includegraphics[width=0.8\linewidth]{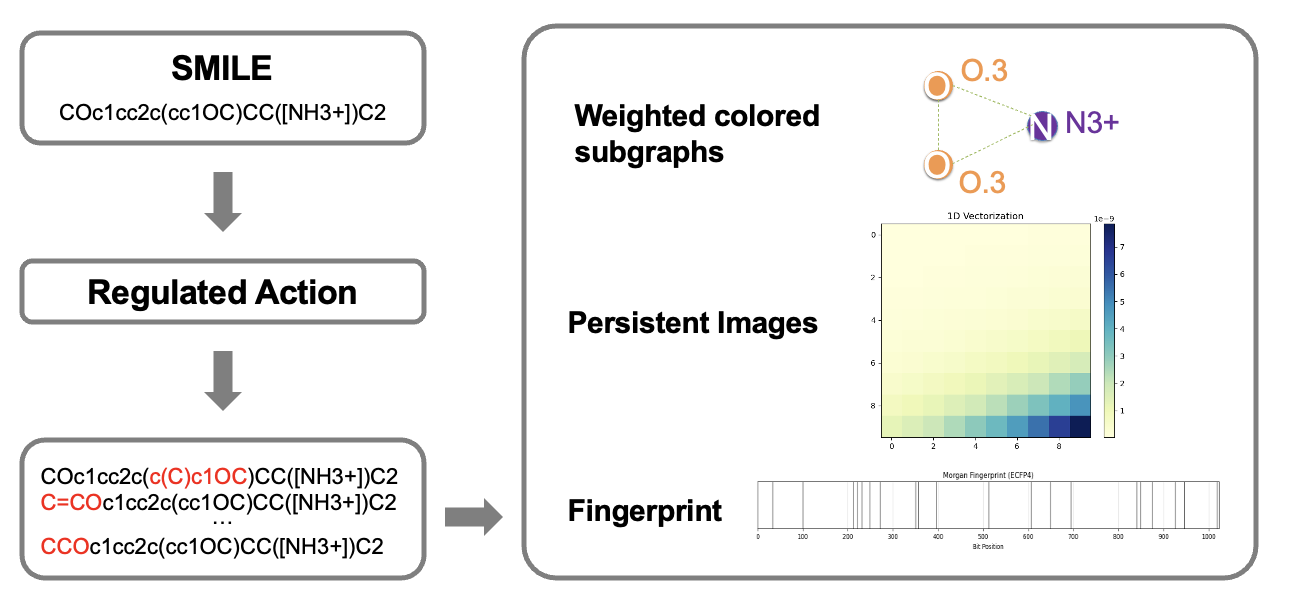}
    \caption{State Construction}
    \label{workflow}
\end{figure}

For each SMILE, the Markov decision process (MDP) is denoted as MDP($\mathcal{S}, \mathcal{A}, {P_{sa}}, \mathcal{R} $). $\mathcal{S}$ denotes the state space, the concatenation of multiscale weighted colored graph feature, persistent image (in the form of a feature vector), and molecular fingerprint. $\mathcal{A}$ denotes the actions space, in which each action $a$ represent the modifications given SMILE. Actions fall into the following three categories: atom addition, bond addition, and bond removal (Fig \ref{action}). ${P_{sa}}$ indicates transition probability and $\mathcal{R}$ indicates reward function. Fig \ref{workflow} shows the flowchart of state construction. 

\subsection{Q-value Function}

We utilize a duel deep neural network to approximate the Q function, $Q(s_i, a_i)$. The dueling network structure separates the estimation of the state value $V$ and the advantage for each action, expressed as $Q\left(s_i, a_i\right)=V\left(s_i\right)+A\left(s_i, a_i\right)$. This architecture allows the model to learn which states are valuable without needing to evaluate every action for each state \cite{wu2023value}, significantly reducing training time.

\subsection{Reward Functions}
The reward functions are tailored to specific optimization tasks, incorporating various molecular properties and constraints.

\subsubsection{Constrained Reward Function}
We integrate Penalized logP and topological complexity into our reward function, properties commonly used to evaluate molecule optimization models \cite{jin2018junction}. The logarithm of the octanol-water partition coefficient, $\log P$, quantifies lipophilicity \cite{maziarka2020mol}. For a molecule $m$, Penalized $\log P(m)$ is defined as:

\begin{equation}
\text{Penalized } \log P(m) = \log P(m) - SA(m)
\end{equation}

where $SA(m)$ denotes the synthetic accessibility of molecule $m$.
The constrained reward function $\mathcal{R}_1$ aims to improve the Penalized logP while maintaining the topological complexity $\mathcal{T}$ (Betti number) and Tanimoto similarity within a specified range, which is achieved by introducing thresholds $\epsilon$, $\delta$ between the optimized molecule $m$ and the original molecule $m_0$:

\begin{align}
R_1(m) &= \text{Penalized } \log P(m) - \lambda \cdot \Big[
\mathbb{I}\left\{S\left(m, m_0\right)<\delta\right\}\left(\delta-S\left(m, m_0\right)\right) \nonumber \\
&\quad + \mathbb{I}\left\{\mathcal{B}\left(m, m_0\right)<\varepsilon\right\}\left(\varepsilon-\mathcal{B}\left(m, m_0\right)\right)
\Big]
\end{align}

Here, $\lambda$ is the coefficient balancing similarity, Penalized logP, and topological complexity, while $\delta$ ensures that $S(m, m_0) \geq \delta$. This constraint is crucial in drug design, as it preserves the core structure or key functional groups often linked to a molecule's biological activity. Additionally, this constraint prevents the model from exploring biologically irrelevant chemical space.

\subsubsection{Target Reward Function}
For tasks involving the design of new molecules with specific molecular weight or Betti number, we define the reward function as:
$$
\mathcal{R}_2(m) = (1 - w) \cdot \mathcal{B}(m) + w \cdot S(m,m_0)
$$
where $\mathcal{B}(m)$ represents the topological feature (e.g., Betti number) of molecule $m$, and $S(m,m_0)$ is the similarity to the original molecule. The coefficient $w$ balances the importance of topological features and molecular similarity.

\section{Experiment}

We utilized 800 molecules randomly selected from the ZINC dataset \cite{irwin2012zinc}. Inspired by the experimental setup of \cite{jin2018junction} and \cite{zhou2019optimization}, we constructed the initial state of the model by combining one of the 800 randomly selected molecules with the \textit{lowest} Penalized logP value and its corresponding MWCG graph feature and persistent image vector. The trained model was then run on each molecule for a single episode. The average improvement in Penalized logP and QED value are reported in the table~\ref{tab:molecule-optimization}.

% \begin{table}[h]
% \centering
% \caption{Top three unique molecule property scores found by each method.}
% \label{tab:molecule-optimization}
% \begin{tabular}{|c|c|c|c|c|c|c|c|c|}
% \hline
% \multirow{2}{*}{Method} & \multicolumn{4}{c|}{Penalized $\log P$} & \multicolumn{4}{c|}{QED} \\
% \cline{2-9}
% & 1st & 2nd & 3rd & Validity & 1st & 2nd & 3rd & Validity \\
% \hline
% random walk & -3.99 & -4.31 & -4.37 & 100\% & 0.64 & 0.56 & 0.56 & 100\% \\
% \hline
% $\varepsilon$-greedy, $\varepsilon=0.1$ & 11.64 & 11.40 & 11.40 & 100\% & 0.914 & 0.910 & 0.906 & 100\% \\
% \hline
% JT-VAE & 5.30 & 4.93 & 4.49 & 100\% & 0.925 & 0.911 & 0.910 & 100\% \\
% \hline
% ORGAN & 3.63 & 3.49 & 3.44 & 0.4\% & 0.896 & 0.824 & 0.820 & 2.2\% \\
% \hline
% GCPN & 7.98 & 7.85 & 7.80 & 100\% & 0.948 & 0.947 & 0.946 & 100\% \\
% \hline
% MolDQN-naïve & 11.51 & 11.51 & 11.50 & 100\% & 0.934 & 0.931 & 0.930 & 100\% \\
% \hline
% MolDQN-bootstrap & 11.84 & 11.84 & 11.82 & 100\% & 0.948 & 0.944 & \textbf{0.943} & 100\% \\
% \hline
% GraphTRL & \textbf{11.89} & \textbf{11.87} & \textbf{11.86} & 100\% & \textbf{0.951} & \textbf{0.949} & \textbf{0.943} & 100\% \\
% \hline
% \end{tabular}
% \end{table}

\begin{table}[h]
\centering
\caption{Top three unique molecule property scores found by each method.}
\label{tab:molecule-optimization}
\begin{tabular}{|c|c|c|c|c|c|c|c|c|}
\hline
\multirow{2}{*}{Method} & \multicolumn{4}{c|}{Penalized $\log P$} & \multicolumn{4}{c|}{QED} \\
\cline{2-9}
& 1st & 2nd & 3rd & Validity & 1st & 2nd & 3rd & Validity \\
\hline
random walk & -3.99 & -4.31 & -4.37 & 100\% & 0.64 & 0.56 & 0.56 & 100\% \\
\hline
$\varepsilon$-greedy, $\varepsilon=0.1$ & 11.64 & 11.40 & 11.40 & 100\% & 0.914 & 0.910 & 0.906 & 100\% \\
\hline
JT-VAE & 5.30 & 4.93 & 4.49 & 100\% & 0.925 & 0.911 & 0.910 & 100\% \\
\hline
ORGAN & 3.63 & 3.49 & 3.44 & 0.4\% & 0.896 & 0.824 & 0.820 & 2.2\% \\
\hline
GCPN & 7.98 & 7.85 & 7.80 & 100\% & 0.948 & 0.947 & 0.946 & 100\% \\
\hline
MolDQN-naïve & 11.51 & 11.51 & 11.50 & 100\% & 0.934 & 0.931 & 0.930 & 100\% \\
\hline
MolDQN-bootstrap & 11.84 & 11.84 & 11.82 & 100\% & 0.948 & 0.944 & 0.930 & 100\% \\
\hline
MolDQN-twosteps & - & - & - & - & 0.948 & 0.948 & \textcolor{red}{0.948} & 100\% \\
\hline
GraphTRL & \textcolor{red}{11.89} & \textcolor{red}{11.87} & \textcolor{red}{11.86} & 100\% & \textcolor{red}{0.951} & \textcolor{red}{0.949} & 0.943 & 100\% \\
\hline
\end{tabular}
\end{table}

\section{Discussion}

This study demonstrates that GraphTRL (Graph-based Topological Reinforcement Learning) offers a flexible approach to molecular optimization and generation, particularly for tasks with specific structural requirements. The model's ability to optimize molecules for various objectives, such as improved Penalized logP or specific Betti numbers, showcases its potential in drug discovery and materials science. By integrating topological features, GraphTRL achieves precise control over molecular structural characteristics, crucial for designing molecules for specific binding sites or creating materials with desired properties. 

% \subsection{Bibliographies}

% Use an unnumbered first-level heading for the references.
% For a citation, use \verb+\cite+, e.g.,~\cite{Kipf:2017tc}.
% For a textual citation, use \verb+\citet+, e.g.,~\citet{Velickovic:2018we}.
% \emph{Any choice} of citation style is allowed as long as it is used consistently throughout the whole paper.
% Additionally, both \verb+natbib+ and \verb+bibLaTeX+ packages are supported.
% It is also possible to reduce the font size to \verb+\small+ (9-point font) when listing the references.

% \section*{Author Contributions}
% Authors of accepted papers are \emph{encouraged} to include a statement that declares the individual contribution of every author, especially when there are co-authors that made equal contributions to the research.
% You may adopt the \href{https://credit.niso.org/}{Contributor Roles Taxonomy (CRediT)} methodology for attributing contributions.
% Do not include this section in the version for blind review.
% This section does not count towards the page limit.

% \section*{Acknowledgements}

% The \LaTeX{} template is heavily borrowed from LoG 2022.

% The acknowledgements do not count towards the page limit.

% For natbib users:
\bibliographystyle{unsrtnat}

% For bibLaTeX users:
% \printbibliography

% \appendix
% \section{Appendix}
% Any possible appendices should be placed after bibliographies.
% If your paper has appendices, please submit the appendices together with the main body of the paper.
% There will be no separate supplementary material submission.
% The main text should be self-contained; reviewers are not obliged to look at the appendices when writing their review comments.

\end{document}